\documentclass[runningheads]{llncs}
\usepackage{graphicx}

\usepackage{hyperref}
\usepackage{subcaption}
\usepackage{adjustbox}
\usepackage{amsmath,amssymb} 
\usepackage{booktabs}
\usepackage{color}
\usepackage[width=122mm,left=12mm,paperwidth=146mm,height=193mm,top=12mm,paperheight=217mm]{geometry}
\usepackage{comment}
\usepackage{makecell}
\usepackage{amssymb}
\usepackage{pifont}
\newcommand{\cmark}{\ding{51}}
\newcommand{\xmark}{\ding{55}}

\begin{document}
\mainmatter

\newcommand\eg{{\emph{e.g.}}}
\newcommand\ie{{\emph{i.e.}}}
\newcommand\etal{{\emph{et al. }}}

\title{Unsupervised Learning of Visual Representations by Solving Jigsaw Puzzles}

\titlerunning{Unsupervised Learning of Visual Representations by Solving Jigsaw Puzzles}

\authorrunning{M. Noroozi and P. Favaro}

\author{Mehdi Noroozi and Paolo Favaro }
\institute{Institute for Informatiks \newline University of Bern \newline \texttt{\{noroozi,paolo.favaro\}@inf.unibe.ch} }

\maketitle

\begin{abstract} 
In this paper we study the problem of image representation learning without human annotation. By following the principles of self-supervision, we build a convolutional neural network (CNN) that can be trained to solve Jigsaw puzzles as a \emph{pretext} task, which requires no manual labeling, and then later repurposed to solve object classification and detection. To maintain the compatibility across tasks we introduce the \emph{context-free network} (CFN), a siamese-ennead CNN. The CFN takes image tiles as input and explicitly limits the receptive field (or context) of its early processing units to one tile at a time. We show that the CFN includes fewer parameters than AlexNet while preserving the same semantic learning capabilities. By training the CFN to solve Jigsaw puzzles, we learn both a feature mapping of object parts as well as their correct spatial arrangement. Our experimental evaluations show that the learned features capture semantically relevant content. Our proposed method for learning visual representations outperforms state of the art methods in several transfer learning benchmarks.
\end{abstract}

\section{Introduction}

Visual tasks, such as object classification and detection, have been successfully approached through the supervised learning paradigm \cite{agrawalGM14,donahueJVHZTD13,KrizhevskyIH12,simonyanZ14}, where one uses labeled data to train a parametric model. However, as manually labeled data can be costly, unsupervised learning methods are gaining momentum.

Recently, Doersch \etal \cite{Carl2015}, Wang and Gupta \cite{Gupta15} and Agrawal \etal \cite{agrawalCM15} have explored a novel paradigm for unsupervised learning called \emph{self-supervised learning}. The main idea is to exploit different labelings that are freely available besides or within visual data, and to use them as intrinsic reward signals to learn general-purpose features. \cite{Carl2015} uses the relative spatial co-location of patches in images as a label. \cite{Gupta15} uses object correspondence obtained through tracking in videos, and \cite{agrawalCM15} uses ego-motion information obtained by a mobile agent such as the Google car \cite{ChenBKTVPRCBPGG11}. The features obtained with these approaches have been successfully transferred to classification and detections tasks, and their performance is very encouraging when compared to features trained in a supervised manner.
\begin{figure}[t]
\centering
\begin{minipage}[c]{.375\textwidth}
\centering
\includegraphics[width=\textwidth,trim=2 2 2 2, clip]{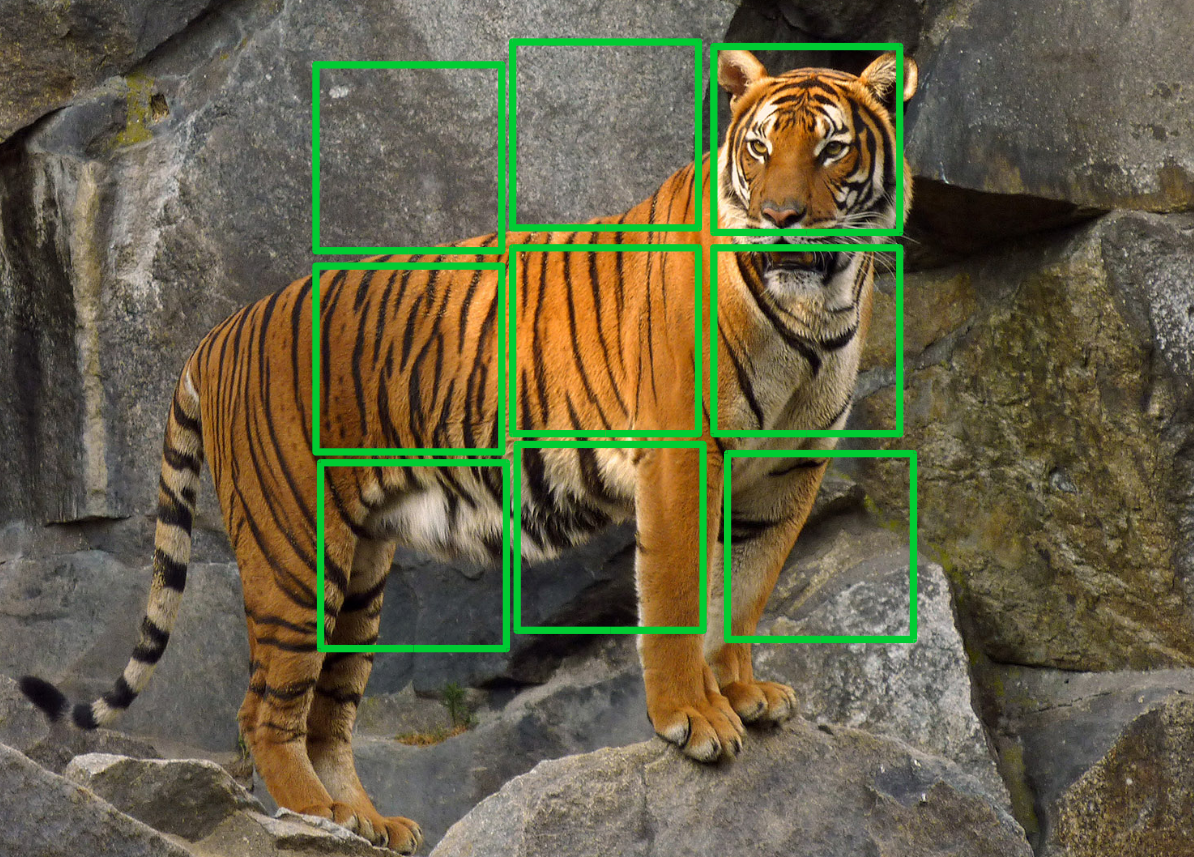}\\
\footnotesize{(a)}
\end{minipage}\hspace{2mm}
\begin{minipage}[c]{.28\textwidth}
\centering
\includegraphics[width=\textwidth,trim=8 8 8 8, clip]{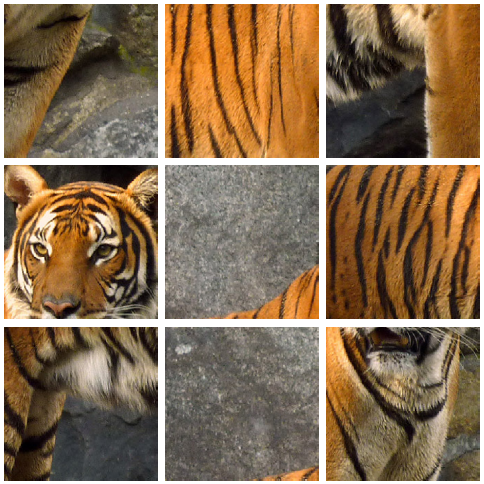}\\
\footnotesize{(b)}
\end{minipage}\hspace{2mm}
\begin{minipage}[c]{.28\textwidth}
\centering
\includegraphics[width=\textwidth,trim=8 8 8 8, clip]{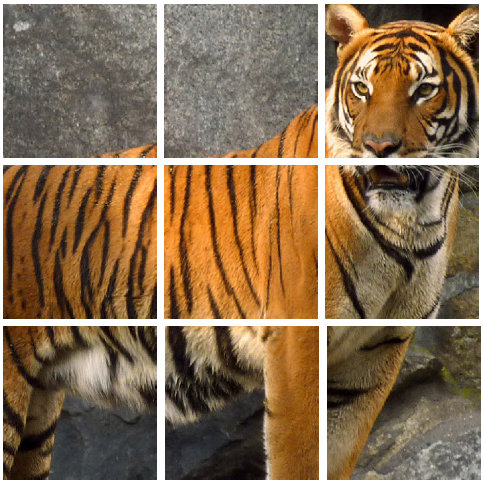}\\
\footnotesize{(c)}
\end{minipage}
 \caption{Learning image representations by solving Jigsaw puzzles. (a) The image from which the tiles (marked with green lines) are extracted. (b) A puzzle obtained by shuffling the tiles. Some tiles might be directly identifiable as object parts, but others are ambiguous (\eg, have similar patterns) and their identification is much more reliable when all tiles are jointly evaluated. 
In contrast, with reference to (c), determining the relative position between the central tile and the top two tiles from the left can be very challenging \cite{Carl2015}.
 \label{fig:puzzle}}
\end{figure}

A fundamental difference between \cite{Carl2015} and \cite{Gupta15,agrawalCM15} is that the former method uses single images as the training set and the other two methods exploit multiple images related either through a temporal or a viewpoint transformation. While it is true that biological agents typically make use of multiple images and also integrate additional sensory information, such as ego-motion, it is also true that single snapshots may carry more information than we have been able to extract so far. This work shows that this is indeed the case. We introduce a novel self-supervised task, the \emph{Jigsaw puzzle reassembly} problem (see Fig.~\ref{fig:puzzle}), which builds features that yield high performance when transferred to detection and classification tasks.

We argue that solving Jigsaw puzzles can be used to teach a system that an object is made of parts and what these parts are. The association of each separate puzzle tile to a precise object part might be ambiguous. However, when all the tiles are observed, the ambiguities might be eliminated more easily because the tile placement is mutually exclusive. This argument is supported by our experimental validation. Training a Jigsaw puzzle solver takes about $2.5$ days compared to $4$ weeks of \cite{Carl2015}. Also, there is no need to handle chromatic aberration or to build robustness to pixelation. Moreover, the features are highly transferrable to detection and classification and yield the highest performance to date for an unsupervised method. 

\section{Related work}

This work falls in the area of \emph{representation/feature learning}, which is an unsupervised learning problem \cite{Barlow89}. Representation learning is concerned with building intermediate representations of data useful to solve machine learning tasks. It also involves \emph{transfer learning} \cite{Yosinski}, as one applies and repurposes features that have been learned by solving the Jigsaw puzzle to other tasks such as object classification and detection.
In our experiments we do so via the \emph{pre-training + fine-tuning} scheme, as in prior work \cite{agrawalCM15}. Pre-training corresponds to the feature learning that we obtain with our Jigsaw puzzle solver. Fine-tuning is instead the process of updating the weights obtained during pre-training to solve another task (object classification or detection).

\paragraph{\textbf{Unsupervised Learning.}} 

There is a rich literature in unsupervised learning of visual representations \cite{bengioCV12}. Most techniques build representations by exploiting general-purpose priors such as smoothness, sharing of factors, factors organized hierarchically, belonging to a low-dimension manifold, temporal and spatial coherence, and sparsity. Unfortunately, a general criterion to design a visual representation is not available. Nonetheless, a natural choice is the goal of disentangling factors of variation. For example, several factors such as the object shapes, the object materials, and the light sources, combine to create complex effects such as shadows, shading, color patterns and reflections in images. Ideal features would separate each of these factors so that other learning tasks (\eg, classification based on just shape or surface materials) can be handled more easily. In this work we design features to separate the appearance from the arrangement (geometry) of parts of objects.

Because of the relevance to contemporary research and to this work, we discuss mainly methods in deep learning. In general one can group unsupervised learning methods into: probabilistic, direct mapping (autoencoders), and manifold learning ones. Probabilistic methods divide variables of a network into observed and latent ones. Learning is then associated with determining model parameters that maximize the likelihood of the latent variables given the observations. A family of popular probabilistic models is the \emph{Restricted Boltzmann Machine} (RBM) \cite{Smolensky86,HintonS86}, which makes training tractable by imposing a bipartite graph between latent and observed variables. 
Unfortunately, these models become intractable when multiple layers are present and are not designed to produce features in an efficient manner. The direct mapping approach focuses on the latter aspect and is typically built via \emph{autoencoders} \cite{BoulardK88,HintonZ94,OlshausenF97}. Autoencoders specify explicitly the feature extraction function (encoder) in a parametric form as well as the mapping from the feature back to the input (decoder). These direct mappings are trained by minimizing the reconstruction error between the input and the output produced by the autoencoder (obtained by applying the encoder and decoder sequentially). A remarkable example of a very large scale autoencoder is the work of Le \etal \cite{LeRMDCCDN12}. Their results showed that robust human and cat faces as well as human body detectors could be built without human labeling.

If the data structure suggests that data points might concentrate around a manifold, then \emph{manifold learning} techniques can be employed \cite{RoweisS2000,BelkinN03}. This representation allows to map directly smooth variations of the factors to smooth variations of the observations. Some of the issues with manifold learning techniques are that they might require computing nearest neighbors (which scales quadratically with the number of samples) and that they need a sufficiently high density of samples around the manifold (and this becomes more difficult to achieve with high-dimensional manifolds).

In the context of Computer Vision, it is worth mentioning some early work on unsupervised learning of models for classification. For instance \cite{Fergus2003,Weber2000} introduced methods to build a probabilistic representation of objects as constellations of parts. A limitation is the high computational complexity of these models. As we will see later, training the Jigsaw puzzle solver also amounts to building a model of both appearance and configuration of the parts.

\paragraph{\textbf{Self-supervised Learning.}} 

This learning strategy is a recent variation on the unsupervised learning theme that exploits labeling that comes for ``free'' with the data \cite{Carl2015,Gupta15,agrawalCM15}. We make a distinction between labels that are easily accessible and are associated with a non-visual signal (for example, ego-motion \cite{agrawalCM15}, but also one could consider audio, text and so on), and labels that are obtained from the structure of the data \cite{Carl2015,Gupta15}. Our work relates to the latter case as we simply re-use the input images and exploit the pixel arrangement as a label. 

Doersch~\etal \cite{Carl2015} train a convolutional network to classify the relative position between two image patches. One tile is kept in the middle of a $3\times 3$ grid and the other tile can be placed in any of the other $8$ available locations (up to some small random shift). In Fig.~\ref{fig:puzzle} (c) we show an example where the relative location between the central tile and the top-left and top-middle tiles is ambiguous. In contrast, the Jigsaw puzzle problem is solved by observing all the tiles at the same time. This allows the trained network to intersect all ambiguity sets and possibly reduce them to a singleton. 

\begin{figure}[t]
\centering
\includegraphics[height=.17\textwidth,trim=10 6 16 14, clip]{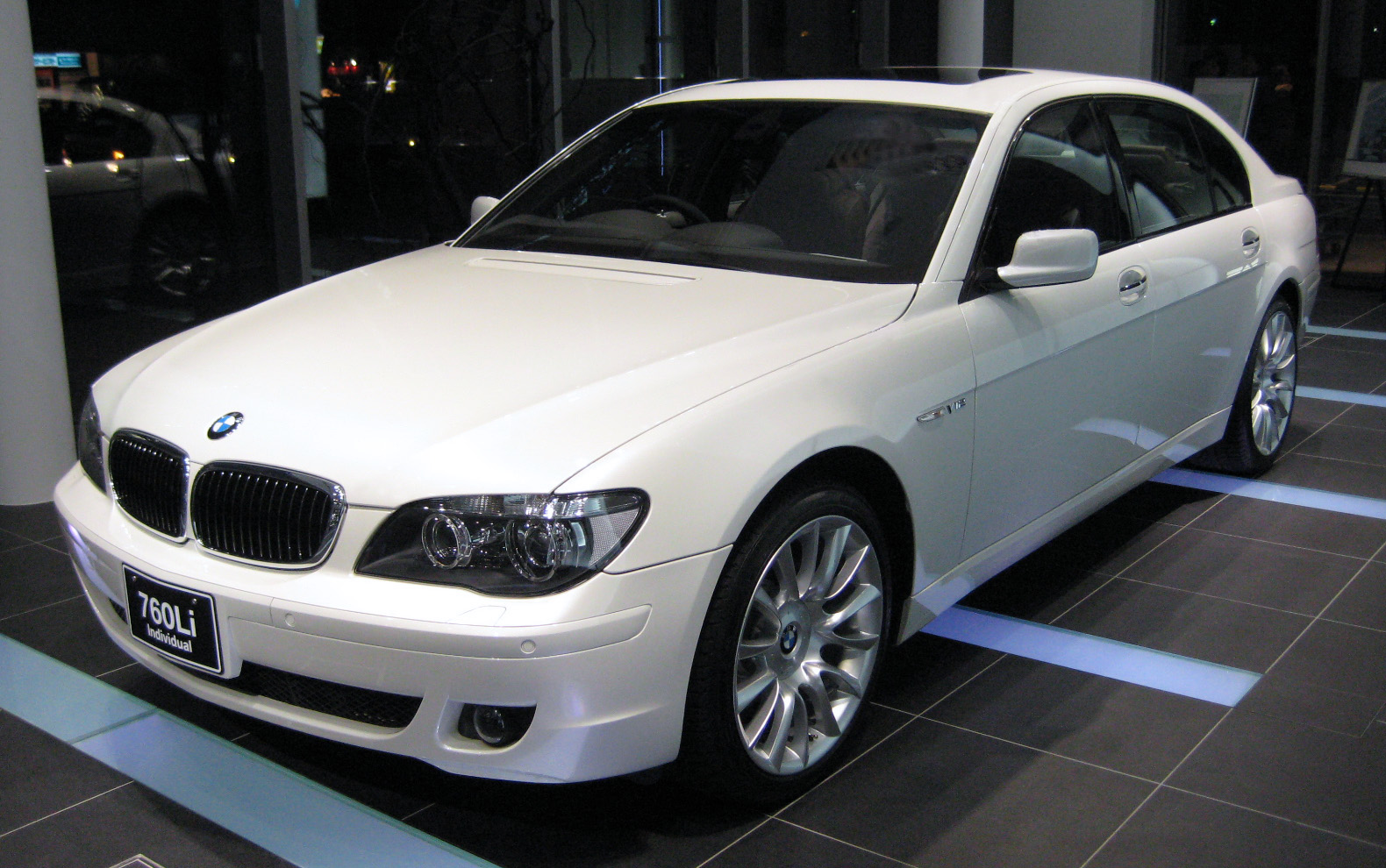}
\includegraphics[height=.17\textwidth,width=.3\textwidth]{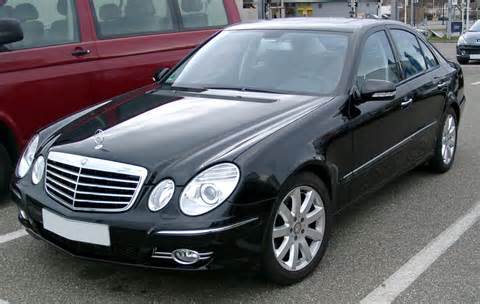}
\includegraphics[height=.17\textwidth,trim=0 35 0 0, clip]{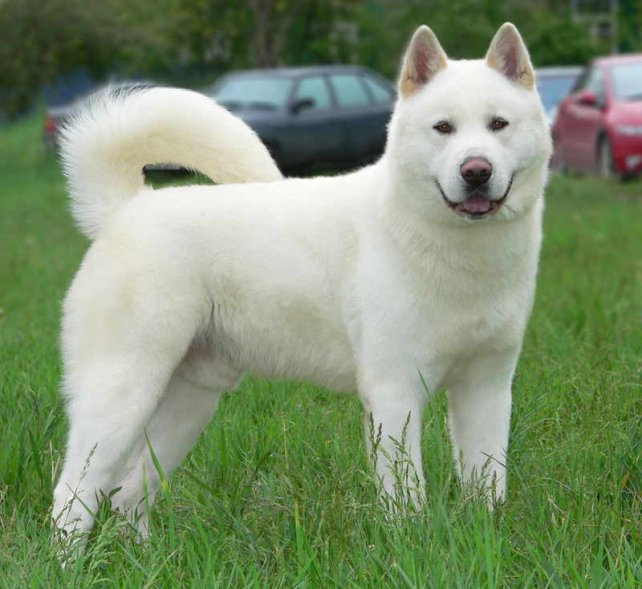}
\includegraphics[height=.17\textwidth,trim=0 35 0 0, clip]{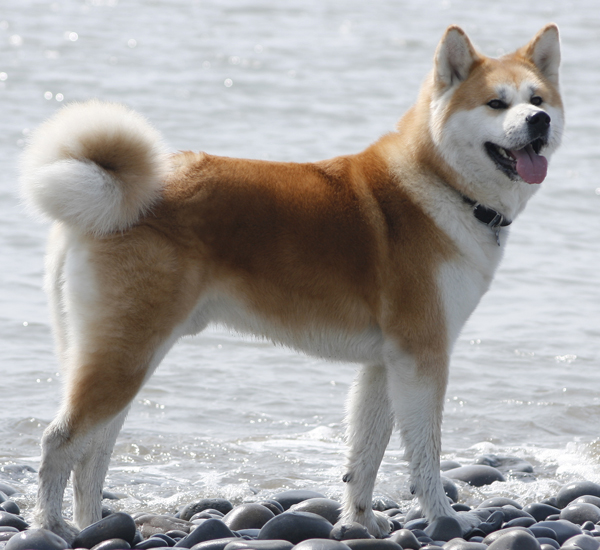}
 \caption{Most of the shape of these $2$ pairs of images is the same (two separate instances within the same categories). However, some low-level statistics are different (color and texture). The Jigsaw puzzle solver learns to ignore such statistics when they do not help the localization of parts.
 \label{fig:parts}}
\end{figure}

The method of Wang and Gupta \cite{Gupta15} builds a metric to define similarity between patches. Three patches are used as input, where two patches are matched via tracking in a video and the third one is arbitrarily chosen. The main advantage of this method is that labeling requires just using a tracking method (they use SURF interest points to detect initial bounding boxes and then tracking via the KCF method \cite{HenriquesCMB15}). The matched patches will have intraclass variability due to changes in illumination, occlusion, viewpoint, pose, occlusions, and clutter factors. However, because the underlying object is the same, the estimated features may not necessarily cluster patches with two different instances of the same object (\ie, based on their semantic content).
The method proposed by Agrawal \etal \cite{agrawalCM15} exploits labeling (egomotion) provided by other sensors. The advantage is that this labeling is freely available in most cases or is quite easy to obtain. They show that egomotion is a useful supervisory signal when learning features. They train a siamese network to estimate egomotion from two image frames and compare it to the egomotion measured with odometry sensors. The trained features will build an invariance similar to that of \cite{Gupta15}. However, because the object identity is the same in both images, the intraclass variability may be limited. With two images of the same instance, learned features focus on their similarities (such as color and texture) rather than their high-level structure. 
In contrast, the Jigsaw puzzle approach ignores similarities between tiles (such as color and texture), as they do not help their localization, and focuses instead on their differences. In Fig.~\ref{fig:parts} we illustrate this concept with two examples: Two cars that have different colors and two dogs with different fur patterns. The features learned to solve puzzles in one (car/dog) image will apply also to the other (car/dog) image as they will be invariant to shared patterns. The ability of the Jigsaw puzzle solver to cluster together object parts can be seen in the top 16 activations shown in Fig.~\ref{fig:activations} and in the image retrieval samples in Fig.~\ref{fig:retrievalquali}.

\paragraph{\textbf{Jigsaw Puzzles.}}

Jigsaw puzzles have been associated with learning since their inception. They were introduced in 1760 by John Spilsbury as a pretext to help children learn geography. The first puzzle was a map attached to a wooden board, which was then sawed in pieces corresponding to countries \cite{Tybon04}. Studies in Psychonomic show that Jigsaw puzzles can be used to assess visuospatial processing in humans \cite{richardsonVecchi02}. Indeed, the \emph{Hooper Visual Organization Test} \cite{Hooper1983} is routinely used to measures an individual's ability to organize visual stimuli. This test uses puzzles with line drawings of simple objects and requires the patient to recognize the object without moving the tiles. Instead of using Jigsaw puzzles to assess someone's visuospatial processing ability, in this paper we propose to use Jigsaw puzzles to develop a visuospatial representation of objects in the context of CNNs.

There is also a sizeable literature on solving Jigsaw puzzles computationally (see, for example,  \cite{Pomeranz04,FreemanG64,PomeranzSB11}). However, these methods rely on the shape of the tiles or on texture especially in the proximity of the borders of the tiles. These are cues that we avoid when training the Jigsaw puzzle solver, as they do not carry useful information when learning a part detector. 

 
\section{Solving Jigsaw Puzzles}
 
At the present time, the design of convolutional neural networks (CNN) is still an art that relies on extensive experience. Here we provide a brief discussion of how we arrived at a convolutional architecture capable of solving Jigsaw puzzles while learning general-purpose features.

An immediate approach to solve Jigsaw puzzles is to stack the tiles of the puzzle along the channels (\ie, the input data would have $9\times 3 = 27$ channels) and then correspondingly increase the depth of the filters of the first convolutional layer in AlexNet \cite{KrizhevskyIH12}. The problem with this design is that the network prefers to identify correlations between low-level texture statistics across tiles rather than between the high-level primitives. Low-level statistics, such as similar structural patterns and texture close to the boundaries of the tile, are simple cues that humans actually use to solve Jigsaw puzzles. However, solving a Jigsaw puzzle based on these cues does not require any understanding of the global object. Thus, here we present a network that delays the computation of statistics across different tiles (see Fig.~\ref{fig:cnnpuzzle}). The network first computes features based only on the pixels within each tile (one row in Fig.~\ref{fig:cnnpuzzle}). Then, it finds the parts arrangement just by using these features (last fully connected layers in Fig.~\ref{fig:cnnpuzzle}). The objective is to force the network to learn features that are as representative and discriminative as possible of each object part for the purpose of determining their relative location.

\subsection{The Context-Free Architecture}

\begin{figure}[t!]
\centering
\includegraphics[width=1\textwidth]{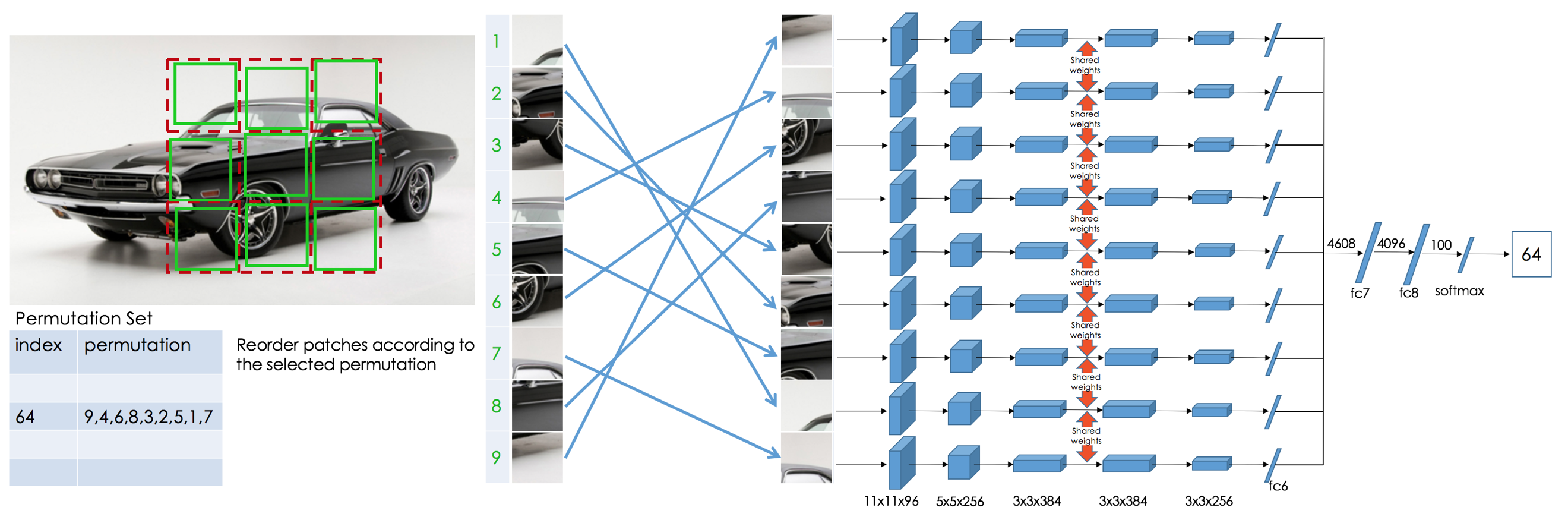}
 \caption{ \textbf{Context Free Network.} The figure illustrates how a puzzle is generated and solved. We randomly crop a $225\times 225$ pixel window from an image (red dashed box), divide it into a $3\times 3$ grid, and randomly pick a $64\times 64$ pixel tiles from each $75\times 75$ pixel cell. These 9 tiles are reordered via a randomly chosen permutation from a predefined permutation set and are then fed to the CFN. The task is to predict the index of the chosen permutation (technically, we define as output a probability vector with $1$ at the $64$-th location and $0$ elsewhere). The CFN is a siamese-ennead CNN. For simplicity, we do not indicate the max-pooling and ReLU layers. These shared layers are implemented exactly as in AlexNet \cite{KrizhevskyIH12}. \textbf{In the transfer learning experiments we show results with the trained weights transferred on AlexNet (precisely, stride 4 on the first layer). The training in the transfer learning experiment is the same as in the other competing methods. Notice instead, that during the training on the puzzle task, we set the stride of the first layer of the CFN to $2$ instead of $4$.}\label{fig:cnnpuzzle}}
\end{figure}

We build a siamese-ennead convolutional network (see\footnote{In earlier versions of this publication we reported transfer learning results where AlexNet had a stride 2 in the first convolutional layer as used during the training for the puzzle task. This arXiv version introduces new updated results. See Fig.~\ref{fig:cnnpuzzle} caption for more information and the Experiments section.} Fig.~\ref{fig:cnnpuzzle}), where each row up to the first fully connected layer (\emph{fc6}) uses the AlexNet architecture \cite{KrizhevskyIH12} with shared weights. Similar schemes were used in prior work \cite{Carl2015,Gupta15,agrawalCM15}. The outputs of all \emph{fc6} layers are concatenated and given as input to $fc7$. All the layers in the rows share the same weights up to and including \emph{fc6}.

We call this architecture the \emph{context-free network} (CFN) because the data flow of each patch is explicitly separated until the fully connected layer and context is handled only in the last fully connected layers. We verify that this architecture performs as well as AlexNet in the classification task on the ImageNet 2012 dataset \cite{dengDSLLF09}. In this test we resize the input images to $225\times 225$ pixels, split them into a $3\times 3$ grid and then feed the full $75\times 75$ tiles to the network.
We find that the CFN achieves $57.1\%$ top-1 accuracy while AlexNet achieves $57.4\%$ top-1 accuracy. However, the CFN architecture is more compact than AlexNet. It depends on only $27.5$M parameters, while AlexNet uses $61$M parameters. The \emph{fc6} layer includes $4\times 4 \times 256 \times 512 \sim 2$M parameters while the \emph{fc6} layer of AlexNet includes $6 \times 6 \times 256 \times 4096 \sim 37.5$M  parameters. However,  the \emph{fc7} layer in our architecture includes $2$M parameters more than the same layer in AlexNet.

This network can thus be used interchangeably for different tasks including detection and classification. In the next section we show how to train the CFN for the Jigsaw puzzle reassembly.

\subsection{The Jigsaw Puzzle Task} 

To train the CFN we define a set of Jigsaw puzzle permutations, \eg, a tile configuration $S = (3,1,2,9,5,4,8,7,6)$, and assign an index to each entry. We randomly pick one such permutation, rearrange the $9$ input patches according to that permutation, and ask the CFN to return a vector with the probability value for each index. Given $9$ tiles, there are $9! = 362{,}880$ possible permutations. From our experimental validation, we found that the permutation set is an important factor on the performance of the representation that the network learns. We perform an ablation study on the impact of the permutation set in subsection~\ref{subsec:perm_set}.

\subsection{Training the CFN}

The output of the CFN can be seen as the conditional probability density function (pdf) of the spatial arrangement of object parts (or scene parts) in a part-based model, \ie,
\begin{align}
p(S|A_1,A_2,\dots,A_9) =  p(S|F_1,F_2,\dots,F_9) \prod_{i=1}^9 p(F_i|A_i)
\end{align}
where $S$ is the configuration of the tiles, $A_i$ is the $i$-th part appearance of the object, and $\{F_i\}_{i=1,\dots,9}$ form the intermediate feature representation.
Our objective is to train the CFN so that the features $F_i$ have semantic attributes that can identify the relative position between parts.

Given the limited amount of data that we can use to build an approximation of this very high-dimensional pdf, close attention must be paid to the training strategy. One problem is when the CFN learns to associate each appearance $A_i$ to an absolute position. \textbf{In this case, the features $F_i$ would carry no semantic meaning, but just information about an arbitrary 2D position}. This problem could happen if we generate just $1$ Jigsaw puzzle per image. Then, the CFN could learn to cluster patches only based on their absolute position in the puzzle, and not on their textural/structural content. If we write the configuration $S$ as a list of tile positions $S=(L_1,\dots,L_9)$ then in this case the conditional pdf $p(S|F_1,F_2,\dots,F_9)$ would factorize into independent terms
\begin{align}
p(L_1,\dots,L_9|F_1,F_2,\dots,F_9) = \prod_{i=1}^9 p(L_i|F_i)
\end{align}
where each tile location $L_i$ is fully determined by the corresponding feature $F_i$. 

More in general, a self-supervised learning system might lead to representations that are suitable to solve the pre-text task, but not the target tasks, \eg, object classification, detection, and segmentation. In this regard, an important factor to learn better representations is to prevent our model from taking these undesirable solutions, such as the one just described above, to solve the pre-text task. We call these solutions \emph{shortcuts}.
Other shortcuts that the model can use to solve the Jigsaw puzzle task include exploiting low-level statistics, such as edge continuity, the pixel intensity/color distribution, and chromatic aberration. 

To avoid shortcuts we employ multiple techniques. To prevent mapping the appearance to an absolute position we feed multiple Jigsaw puzzles of the same image to the CFN (an average of $69$ out of $1000$ possible puzzle configurations) and make sure that the tiles are shuffled as much as possible by choosing configurations with sufficiently large average Hamming distance. In this way the same tile would have to be assigned to multiple positions (possibly all $9$) thus making the mapping of features $F_i$ to any absolute position equally likely. 
To avoid shortcuts due to edge continuity and pixel intensity distribution we also leave a random gap between the tiles. This discourages the CFN from learning low-level statistics and was also done in \cite{Carl2015}. During training we resize each input image until either the height or the width matches $256$ pixels and preserve the original aspect ratio. Then, we crop a random region from the resized image of size $225\times225$ and split it into a $3\times 3$ grid of $75\times 75$ pixels tiles. We then extract a $64\times 64$ region from each tile by introducing random shifts and feed them to the network. Thus, we have an average gap of $11$ pixels between the tiles. However, the gaps may range from a minimum of $0$ pixels to a maximum of $22$ pixels.
To avoid shortcuts due to chromatic aberration we jitter the color channels and use grayscale images (see more details in the Experiments section). 
In subsection~\ref{subsec:shortcuts} we perform ablation studies on the techniques we use to prevent the shortcuts. 

We used Caffe \cite{jia2014caffe} and modified the code to choose random image patches and permutations during the training time. This allowed us to keep the dataset small ($1.3$M images from ImageNet) and the training efficient, while the CFN could see an average of $69$ different puzzles per image (that is about $90$M different Jigsaw puzzles).

\subsection{Implementation Details}

We use stochastic gradient descent without batch normalization \cite{batch_normalization} on one Titan X GPU. The training uses $1.3$M color images of $256\times 256$ pixels and mini-batches with a batch size of $256$ images. The images are resized by preserving the aspect ratio until either the height or the width matches $256$ pixels. Then the other dimension is cropped to $256$ pixels. The training converges after $350$K iterations with a basic learning rate of $0.01$ and takes $59.5$ hours in total ($\sim 2.5$ days). If we take $122\%= \frac{3072\text{cores}@1000\text{Mhz}}{2880 \text{cores} @875\text{Mhz}}=\frac{6,144\text{GFLOPS}}{5,040\text{GFLOPS}}$ as the best possible performance ratio between the Titan X and the Tesla K40 (used for \cite{Carl2015}) we can predict that the CFN would have taken $\sim 72.5$ hours ($\sim 3$ days) on a Tesla K40.
We compute that on average each image is used $350K \times 256 / 1.3M \simeq 69$ times. That is, we solve on average $69$ Jigsaw puzzles per image.

\section{Experiments}
We first evaluate the performance of our learned representations on different transfer learning benchmarks. We then perform ablation studies on our proposed method. We also visualize the neurons of the intermediate layers of our network. Finally, we compare our features with those of \cite{Carl2015,Gupta15} both qualitatively and quantitatively on image retrieval.

\subsection{Transfer Learning}
We evaluate our learned features as pre-trained weights for classification, detection, and semantic segmentation tasks on the PASCAL VOC dataset\cite{voc}.
We also introduce a novel benchmark to evaluate methods for unsupervised/self-supervised representation learning. 
After training the CFN on the self-supervised learning task, we use the CFN weights to initialize all the \texttt{conv} layers of a standard AlexNet network (stride 4 on the first layer). Then, we retrain the rest of the network from scratch (Gaussian noise as initial weights) for object classification on ImageNet dataset. Notice that while during the training of the Jigsaw task we use stride 2 in the first layer of our CFN, we use a standard AlexNet (stride 4 on the first layer) to make the comparison with competing methods in all the experiments directly comparable. 

\subsubsection{Pascal VOC}
We fine-tune the Jigsaw task features on the classification task on PASCAL VOC 2007 by using the framework of Kr\"ahenb\"uhl \etal \cite{kraehenbuehlDDD16} and on the object detection task by using the Fast R-CNN \cite{fastrcnn} framework. We also fine-tune our weights for the semantic segmentation task using the framework \cite{fcn} on the PASCAL VOC 2012 dataset. Because our fully connected layers are different from those of the standard AlexNet, we select one row of the CFN (up to \texttt{conv5}), copy only the weights of the convolutional layers, and fill the fully connected layers with Gaussian random weights with mean $0.1$ and standard deviation $0.001$. The results are summarized in Table~\ref{tab:voc_cl_dtn}.

Our features achieve $53.2\%$ mAP using multi-scale training and testing, $67.6\%$ in classification, and $37.6\%$ in semantic segmentation thus outperforming all other methods and closing the gap with features obtained with supervision.

\begin{table}[t]
\caption{Results on PASCAL VOC 2007 Detection and Classification. The results of the other methods are taken from Pathak \etal\cite{context_encoder}. \label{tab:voc_cl_dtn}}
\small
\begin{adjustbox}{max width=\textwidth}
\begin{tabular}{ l@{\hspace{1.5em}}  c@{\hspace{1.1em}} c@{\hspace{1.1em}} c@{\hspace{1.1em}} c@{\hspace{1.1em}} c@{\hspace{1.1em}}}

\toprule
\textbf{Method} & \textbf{Pretraining time} & \textbf{Supervision} & \textbf{Classification} & \textbf{Detection} & \textbf{Segmentation}\\
\midrule
Krizhevsky\etal\cite{KrizhevskyIH12} & 3 days &  1000 class labels & \textbf{78.2\%} & \textbf{56.8\%} &  \textbf{48.0\%}\\
\midrule
Wang and Gupta\cite{Gupta15} & 1 week &   motion & 58.4\% & 44.0\% & - \\
Doersch \etal\cite{Carl2015} & 4 weeks &  context & 55.3\% & 46.6\% & -\\
Pathak \etal\cite{context_encoder} &  14 hours & context & 56.5\% & 44.5\% & 29.7\% \\
Ours & 2.5 days &  context & \textbf{67.6\%} & \textbf{53.2\%} & \textbf{37.6\%} \\
\bottomrule

\end{tabular}
\end{adjustbox}
\end{table}

\subsubsection{ImageNet Classification}

Yosinski~\etal \cite{Yosinski} have shown that the last layers of AlexNet are specific to the task and dataset used for training, while the first layers are general-purpose. In the context of transfer learning, this transition from general-purpose to task-specific determines where in the network one should extract the features. In this section we try to understand where this transition occurs in our learned representation. We repurpose our weights, \cite{Carl2015}, and \cite{Gupta15} to the classification task on the ImageNet 2012 dataset \cite{dengDSLLF09}. Table~\ref{ps2s} summarizes the results. The analysis consists of training each network with the labeled data from ImageNet 2012 by locking a subset of the layers and by initializing the unlocked layers with random values. 
If we train AlexNet, we obtain the reference maximum accuracy of $57.4\%$. Our method achieves $34.6\%$ when only fully connected layers are trained. There is a significant improvement (from $34.6\%$ to $45.3\%$) when the \texttt{conv5} layer is also trained. This shows that the \texttt{conv5} layer starts to be specialized on the Jigsaw puzzle reassembly task.

We also perform a novel experiment to understand whether semantic classification is useful to solve Jigsaw puzzles, and thus to see how much object classification and Jigsaw puzzle reassembly tasks are related. We take the pre-trained AlexNet and transfer its features to solve Jigsaw puzzles. We also use the same locking scheme to see the transferability of features at different layers. The performance is shown in Table~\ref{tab:rectopuzzle}. Compared to the maximum accuracy of the Jigsaw task, $88\%$, we can see that semantic training is quite helpful towards recognizing object parts. Indeed, the performance is very high up to \texttt{conv4}.

\begin{table}[t!]
\caption {Comparison of classification results on ImageNet 2012 \cite{image_net}. The numbers are obtained by averaging $10$ random crops predictions. 
 \label{tab:classification}\label{ps2s}}
\footnotesize
\begin{center}
\begin{tabular}{ l c@{\hspace{1em}} c@{\hspace{1.8em}} c@{\hspace{1.8em}} c@{\hspace{1.8em}} c@{\hspace{1.8em}} c@{\hspace{1.8em}} }
\toprule
   & & \includegraphics[height=.7em]{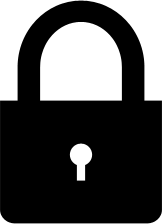} conv1 
     & \includegraphics[height=.7em]{lockmin} conv2 
     & \includegraphics[height=.7em]{lockmin} conv3
     & \includegraphics[height=.7em]{lockmin} conv4
     & \includegraphics[height=.7em]{lockmin} conv5 \\
  \midrule
   CFN & &  \textbf{54.7} & \textbf{52.8} & \textbf{49.7} & 45.3 & \textbf{34.6} \\
   Doersch \etal \cite{Carl2015} & & 53.1 & {47.6} & 48.7 & \textbf{45.6} & 30.4 \\
   Wang and Gupta \cite{Gupta15} & & 51.8 & 46.9 & 42.8 & 38.8 & 29.8 \\
   Random & & 48.5 & 41.0 & 34.8 & 27.1 & 12.0 \\
  \bottomrule
\end{tabular}
\end{center}
\end{table}

\begin{table}[t!]
 \caption{Transfer learning of AlexNet from a classification task to the Jigsaw puzzle reassembly problem. The $j$-th column indicates that all layers from \texttt{conv1} to \texttt{conv-}$j$ were locked and all subsequent layers were randomly initialized and retrained. Notice how the first $4$ layers provide very good features for solving puzzles. This shows that object classification and the Jigsaw puzzle problems are related.\label{tab:rectopuzzle}}
\footnotesize
\begin{center}
\begin{tabular}{ c@{\hspace{2em}} | c@{\hspace{2em}} c@{\hspace{2em}} c@{\hspace{2em}} c@{\hspace{2em}} c@{\hspace{2em}} }
&     \includegraphics[height=1em]{lockmin} conv1
   & \includegraphics[height=1em]{lockmin} conv2 
   & \includegraphics[height=1em]{lockmin} conv3
   & \includegraphics[height=1em]{lockmin} conv4
   & \includegraphics[height=1em]{lockmin} conv5 \\
   \hline
AlexNet \cite{KrizhevskyIH12}&   88 & 87 & 86 & 83 & 74 \\
\end{tabular}
\end{center}
\end{table}

\begin{table}[t]
 \caption{Ablation study on the impact of the permutation set.}
\label{tbl:perm_set}
 
\begin{adjustbox}{width=\textwidth}
\begin{tabular}{c@{\hspace{2em}}  c@{\hspace{2em}}  c@{\hspace{2em}} c@{\hspace{2em}} c}
\toprule
\textbf{Number of }   & \textbf{Average hamming} & \textbf{Minimum hamming} &  \textbf{Jigsaw task }  & \textbf{Detection } \\
\textbf{permutations}   & \textbf{distance} & \textbf{distance} &  \textbf{accuracy}  & \textbf{performance} \\
\midrule
1000 & 8.00 & 2 &  71 & \textbf{53.2}\\
1000 & 6.35 & 2 &  62 & 51.3\\
1000 & 3.99 & 2 &  54 & 50.2\\
\midrule
100 & 8.08 & 2 &  88 & 52.6 \\
95 & 8.08 & 3 & 90 & 52.4   \\
85 & 8.07 & 4 & 91 & 52.7   \\
71 & 8.07 & 5 & 92 & 52.8   \\
35 & 8.13 & 6 & 94 & 52.6   \\
10 & 8.57 & 7 & 97 & 49.2   \\
7  & 8.95 & 8 & 98 & 49.6   \\
6  & 9    & 9 & 99 & 49.7   \\
\bottomrule
\end{tabular}
\end{adjustbox}
\end{table}

\begin{table}[t]
 \caption{Ablation study on the impact of the shortcuts.}
\label{tbl:shortcut}
\begin{adjustbox}{width=\textwidth}
\begin{tabular}{c@{\hspace{2em}} c@{\hspace{2em}} c@{\hspace{2em}} c@{\hspace{2em}} c}
\toprule
\textbf{Gap} &  \textbf{ Normalization} & \textbf{Color jittering}  &  \textbf{Jigsaw task accuracy} & \textbf{Detection performance}\\
\midrule
 \xmark   & \cmark & \cmark & 98 & 47.7 \\
 \cmark & \xmark & \cmark & 90 & 43.5 \\
 \cmark & \cmark & \xmark & 89 & 51.1\\
 \cmark  & \cmark & \cmark & 88 & 52.6 \\
\bottomrule
\end{tabular}
\end{adjustbox}
\end{table}

\begin{table}[t!]
\label{algo:haming}
\begin{tabular}{ l@{\hspace{3em}}}
\Xhline{2\arrayrulewidth}
  \textbf{Algorithm 1.} Generation of the \textit{maximal} Hamming distance permutation set\\
 \hline 
\end{tabular} 
\begin{tabular}{ l@{\hspace{2em}} l}
\textbf{Input:} $N$    &$\backslash\backslash$ \emph{number of permutations} \\
 \textbf{Output:} $P$ &$\backslash\backslash$ \emph{maximal permutation set} \\
 1: $\bar P \leftarrow$ all permutations $[\bar P_1,\dots,\bar P_{9!}]$&$\backslash\backslash$ \emph{$\bar P$ is a $9 \times 9!$ matrix}\\
 2: $P \leftarrow \emptyset$ \\
 3: $j \sim {\cal U}[1,9!]$& \hfill$\backslash\backslash$ \emph{uniform sample out of $9!$ permutations} \\
 4: $i \leftarrow 1$&\\
 5: \textbf{repeat} &\\
 6: $\quad$ $P \leftarrow [P~ \bar P_j]$ & $\backslash\backslash$ \emph{add permutation $\bar P_j$ to $P$} \\
 7: $\quad$ $\bar P \leftarrow  [\bar P_1,\dots,\bar P_{j-1}, \bar P_{j+1}, \dots]$ & $\backslash\backslash$ \emph{remove $\bar P_j$ from $\bar P$} \\
 8: $\quad$ $D \leftarrow \text{Hamming}(P,P^{\prime})$ & $\backslash\backslash$ \emph{$D$ is an $i \times (9!-i)$ matrix}\\
 9: $\quad$ $\bar D \leftarrow \mathbf{1}^T D$ &$\backslash\backslash$ \emph{$\bar D$ is a $1 \times (9!-i)$ row vector}\\
10:$\quad$ $j \leftarrow \arg \max_{k} \bar D_k$ &$\backslash\backslash$ \emph{$\bar D_k$ denotes the $k$-th entry of $\bar D$}\\
11:$\quad$ $i \leftarrow i+1$ &\\
12: \textbf{until} $i \leq N$ &\\
\hline
\end{tabular}
\end{table}

\subsection{Ablation Studies}
We perform ablation studies on our proposed methods to show the impact of each component during the training of Jigsaw task. We train under different scenarios and evaluate the performance on detection task on PASCAL VOC 2007.  

\subsubsection{Permutation Set.} \label{subsec:perm_set} The permutation set controls the ambiguity of the task. If the permutations are close to each other, the Jigsaw puzzle task is more challenging and ambiguous. For example, if the difference between two different permutations lies only in the position of two tiles and there are two similar tiles in the image, the prediction of the right solution will be impossible. The challenge here is a weaker version of what happens in the method of Doersch~\etal~\cite{Carl2015}. To show this issue quantitatively, we compare the performance of the learned representation on the PASCAL VOC 2007 detection task by generating several permutation sets based on the following three criteria:
\paragraph{I) Cardinality.} We train the network with a different number of permutations and see what impact this has on the learned features. We find that as the total number of permutations increases, the training on the Jigsaw task becomes more and more difficult. Also, we find that the performance of the detection task increases with a growing number of permutations. 
\paragraph{II) Average Hamming distance.}   We use a subset of $1000$ permutations and select them based on their Hamming distance (\ie, the number of different tile locations between $2$ permutations $S_1$ and $S_2$). One can see that the average Hamming distance between permutations controls the difficulty of the Jigsaw puzzle reassembly task, and it also correlates with the object detection performance. We find that as the average Hamming distance increases, the CFN yields lower Jigsaw puzzle solving errors and lower object detection errors with fine-tuning. In the Experiments section we compare the performance on object detection of CFNs trained with $3$ choices for the Hamming distance: minimal, average and maximal (see Table~\ref{tbl:perm_set}). From those tests we can see that large Hamming distances are desirable.
We generate this permutation set iteratively via a greedy algorithm. We begin with an empty permutation set and at each iteration select the one that has the desired Hamming distance to the current permutation set. Algorithm 1 provides more details about the algorithm. For the minimal and middle case, the $\arg\max_{k}$ function at line $10$ is replaced by $\arg\min_{k}$ and uniform sampling respectively. Note that the permutation set is generated before training.
\paragraph{III) Minimum hamming distance.} To increase the minimum possible distance between permutations, we remove similar permutations in a maximal set with $100$ initial entries. As argued before, the minimum distance helps to make the task less ambiguous. The performance results showing the impact of each component are summarized in Table~\ref{tbl:perm_set}. The best performing permutation set is a trade off between the number of permutations and how dissimilar they are from each other. 

 \vspace{.5em}
The outcome of this ablation study seems to point to the following final consideration:
 \vspace{-.5em}
 \begin{center}
  \textit{A good self-supervised task is neither simple nor ambiguous.} 
 \end{center}
 
\subsubsection{Preventing Shortcuts} \label{subsec:shortcuts} In a self-supervised learning method, \textit{shortcuts} exploit information useful for solving the pre-text task, but not for a target task, such as detection. 
Similar to \cite{Carl2015}, we experimentally show that the CFN can take the following shortcuts to solve the Jigsaw Puzzle task: 

\paragraph{Low level statistics.} Adjacent patches include similar low-level statistics like the mean and standard deviation of the pixel intensities. This allows the model to find the arrangement of the patches. To avoid this shortcut, we normalize the mean and the standard deviation of each patch independently.
\paragraph{Edge continuity.} A strong cue to solve Jigsaw puzzles is the continuity of edges. We select the $64\times 64$ pixel tiles randomly from the $85 \times 85$ pixel cells. This allows us to have a $21$ pixel gap between tiles.
\paragraph{Chromatic Aberration.} Chromatic aberration is a relative spatial shift between color channels that increases from the images center to the borders. This type of distortion helps the network to estimate the tile positions. To avoid this shortcut, we use three techniques: i) We crop the central square of the original image and resize it to $255 \times 255$; ii) We train the network with both color and grayscale images. Our training set is a composition of grayscale and color images with a ratio of $30\%$ to $70\%$; iii) We (spatially) jitter the color channels of the color images of each tile randomly by ${\pm 0,\pm 1,\pm 2}$ pixels.

 \vspace{.5em}
Table~\ref{tbl:shortcut} shows the performance of transfer learning our CFN, trained under different combinations of the above techniques to avoid shortcuts, to the detection task on Pascal VOC.

\subsection{CFN filter activations} 

\begin{figure}[t!]
\begin{center}
\begin{subfigure}{0.49\textwidth}
\includegraphics[width=1\textwidth,trim=0 0 0 0,clip]{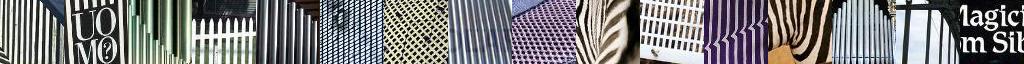}
\includegraphics[width=1\textwidth,trim=0 0 0 0,clip]{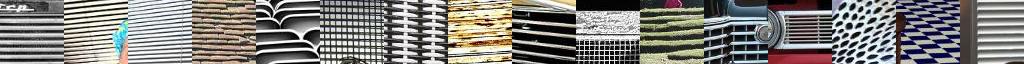}
\includegraphics[width=1\textwidth,trim=0 0 0 0,clip]{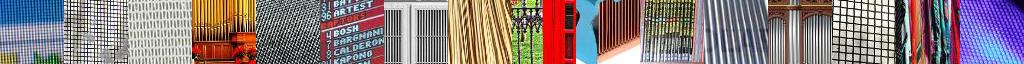}
\includegraphics[width=1\textwidth,trim=0 0 0 0,clip]{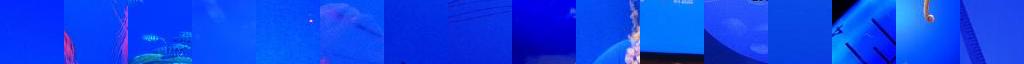}
\includegraphics[width=1\textwidth,trim=0 0 0 0,clip]{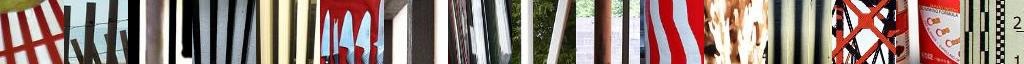}
\includegraphics[width=1\textwidth,trim=0 0 0 0,clip]{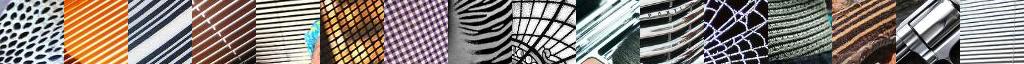}
\caption{conv1 activations}
\end{subfigure}
\begin{subfigure}{0.49\textwidth}
\includegraphics[width=1\textwidth,trim=0 0 0 0,clip]{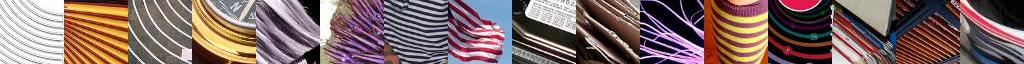}
\includegraphics[width=1\textwidth,trim=0 0 0 0,clip]{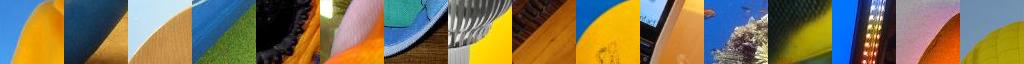}
\includegraphics[width=1\textwidth,trim=0 0 0 0,clip]{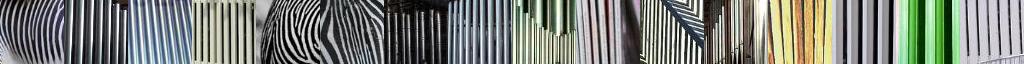}
\includegraphics[width=1\textwidth,trim=0 0 0 0,clip]{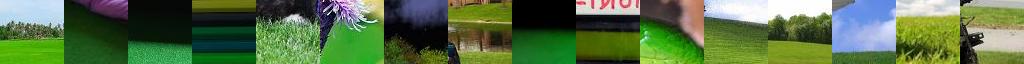}
\includegraphics[width=1\textwidth,trim=0 0 0 0,clip]{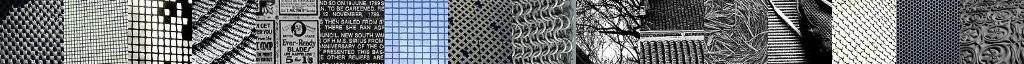}
\includegraphics[width=1\textwidth,trim=0 0 0 0,clip]{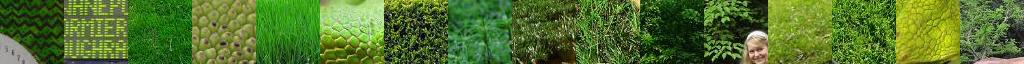}
\caption{conv2 activations}
\end{subfigure}\vspace{1mm}
\begin{subfigure}{0.49\textwidth}
\includegraphics[width=1\textwidth,trim=0 0 0 0,clip]{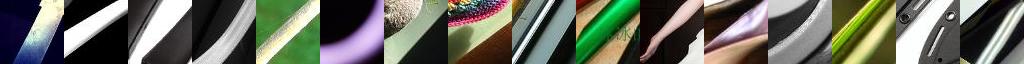}
\includegraphics[width=1\textwidth,trim=0 0 0 0,clip]{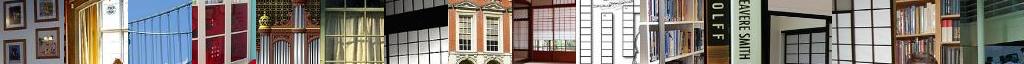}
\includegraphics[width=1\textwidth,trim=0 0 0 0,clip]{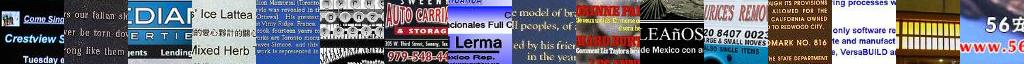}
\includegraphics[width=1\textwidth,trim=0 0 0 0,clip]{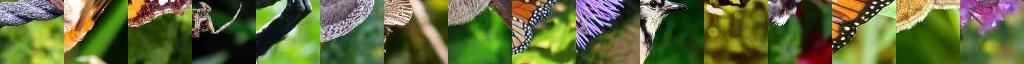}
\includegraphics[width=1\textwidth,trim=0 0 0 0,clip]{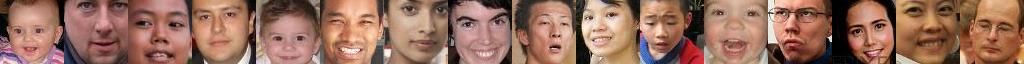}
\includegraphics[width=1\textwidth,trim=0 0 0 0,clip]{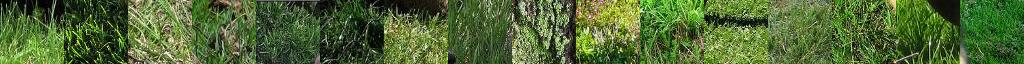}
\caption{conv3 activations}
\end{subfigure}
\begin{subfigure}{0.49\textwidth}
\includegraphics[width=1\textwidth,trim=0 0 0 0,clip]{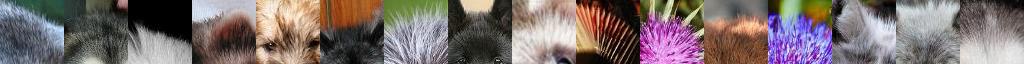}
\includegraphics[width=1\textwidth,trim=0 0 0 0,clip]{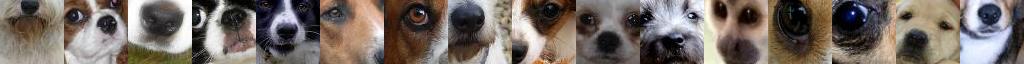}
\includegraphics[width=1\textwidth,trim=0 0 0 0,clip]{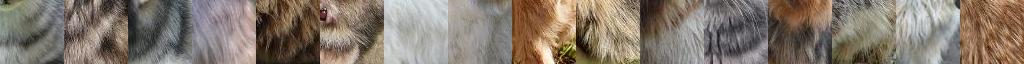}
\includegraphics[width=1\textwidth,trim=0 0 0 0,clip]{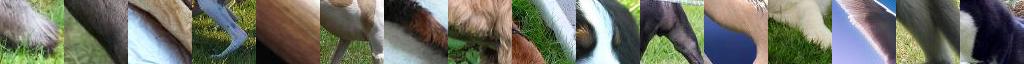}
\includegraphics[width=1\textwidth,trim=0 0 0 0,clip]{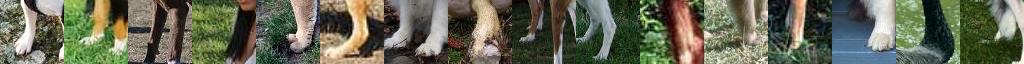}
\includegraphics[width=1\textwidth,trim=0 0 0 0,clip]{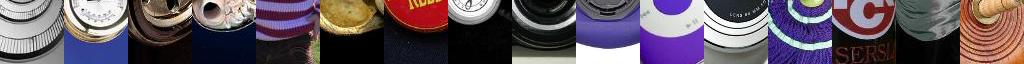}
\caption{conv4 activations}
\end{subfigure}\vspace{1mm}
\begin{subfigure}{0.49\textwidth}
\includegraphics[width=1\textwidth,trim=0 0 0 0,clip]{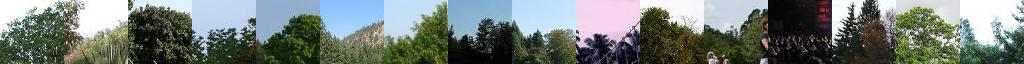}
\includegraphics[width=1\textwidth,trim=0 0 0 0,clip]{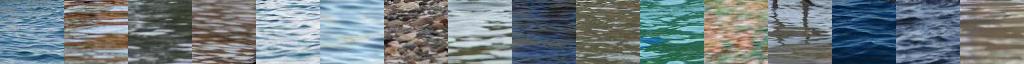}
\includegraphics[width=1\textwidth,trim=0 0 0 0,clip]{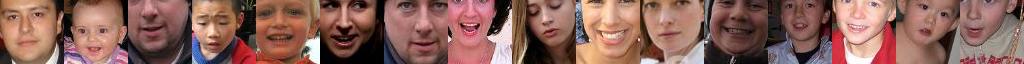}
\includegraphics[width=1\textwidth,trim=0 0 0 0,clip]{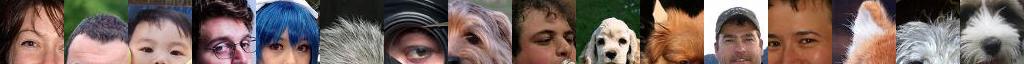}
\includegraphics[width=1\textwidth,trim=0 0 0 0,clip]{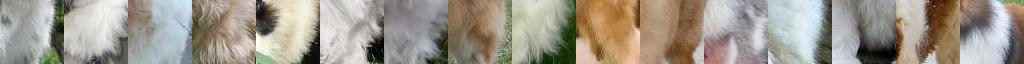}
\includegraphics[width=1\textwidth,trim=0 0 0 0,clip]{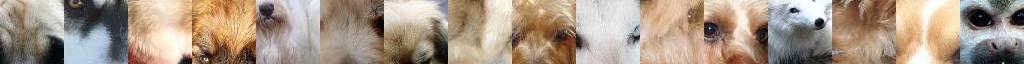}
\caption{conv5 activations}
\end{subfigure}
\begin{subfigure}{0.49\textwidth}
\includegraphics[width=1\textwidth,height=.39\textwidth,trim=0 0 0 0,clip]{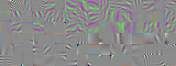}
\caption{conv1 filters without color jittering }
\end{subfigure}
\begin{subfigure}{0.49\textwidth}
\includegraphics[width=1\textwidth,height=.39\textwidth,trim=0 0 0 0,clip]{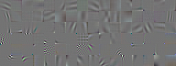}
\caption{conv1 filters with color jittering}
\end{subfigure}

\caption{Visualization of the top $16$ activations for $6$ units of the \texttt{conv1}, \texttt{conv2}, \texttt{conv3}, \texttt{conv4}, \texttt{conv5} layers in our CFN trained without blocking chromatic aberration. (f),(g) we show the filters of \texttt{conv1} trained without and with blocking chromatic aberration. The selection of the top activations is identical to the visualization method of Girshick~\etal \cite{GirshickDDM2014}, except that we compute the average response rather than the maximum. We show some of the most significant units. We can see that in the first (a) and second (b) layers the filters specialize on different types of textures. On the third layer (c) the filters become more specialized and we have a first face detector (later layers will also have face detectors in some units) and some part detectors (\eg, the bottom corner of the butterflies wing). On the fourth layer (d) we have already quite a number of part detectors. We purposefully choose all the dog part detectors: head top, head center, neck, back legs, and front legs. Notice the intraclass variation of the parts. Lastly, the fifth convolutional layer (e) has some other part detectors and some scene part detectors.} \label{fig:activations}
\end{center}
\end{figure}

Some recent work has devoted efforts towards the visualization of CNNs to better understand how they work and how we can exploit them \cite{ZeilerF2014,simonyanVZ14,mahendranV15,YosinskiCNFL2015}. Some of these works 
aim at obtaining the input image that best represents a category according to a given neural network. This has shown that CNNs retain important information about the categories. Here instead we analyze the CFN by considering the units at each layer as object part detectors as in \cite{GirshickDDM2014}. We extract $1$M patches from the ImageNet validation set ($20$ randomly sampled $64\times 64$ patches) and feed them as input to the CFN. At each layer (\texttt{conv1}, \texttt{conv2}, \texttt{conv3}, \texttt{conv4}, \texttt{conv5}) we consider the outputs of one channel and compute their $\ell_1$ norm. We then rank the patches based on the $\ell_1$ norm and select the top $16$ ones that belong to different images. Since each layer has several channels, we hand-pick the $6$ most significant ones. In Fig.~\ref{fig:activations} we show the top-16 activation patches for only $6$ channels per layer. These activations show that the CFN features correspond to patterns sharing similar shapes and that there is a good correspondence based on object parts (in particular see the \texttt{conv4} activations for dog parts). Some channels seem to be good face detectors (see \texttt{conv3}, but the same detectors can be seen in other channels, not shown, in \texttt{conv4} and \texttt{conv5}) and others seem to be good texture detectors (\eg, grass, water, fur). In Fig.~\ref{fig:activations}(f) we also show the filters of the \texttt{conv1} layer of the CFN. We can see that these filters are quite strong and our transfer learning experiments in the next sections show that they are as effective as those trained in a supervised manner.

\subsection{Image Retrieval}  
\begin{figure}[t!]
\begin{center}
\includegraphics[height=.046\textwidth,trim=0 22.6mm 452mm 22.6mm,clip]{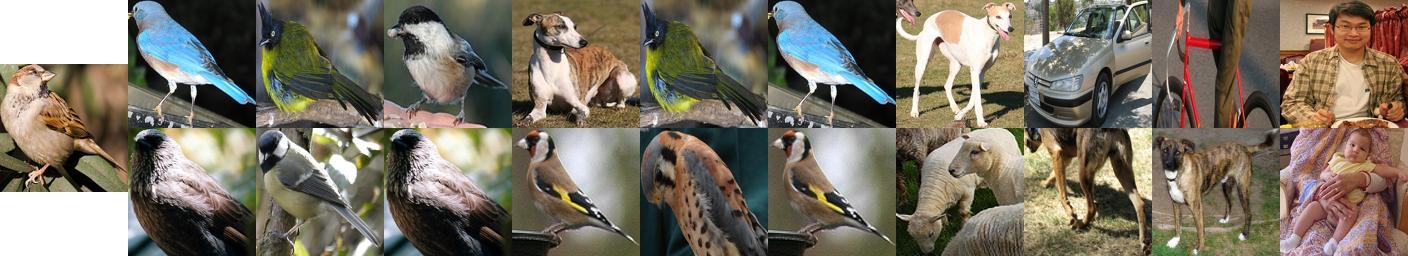}
\includegraphics[height=.046\textwidth,trim=135.6mm 45.2mm 271.2mm 0,clip]{ret_1.jpg}\hspace{-1mm}
\includegraphics[height=.046\textwidth,trim=135.6mm 0 271.2mm 45.2mm,clip]{ret_1.jpg}
\includegraphics[height=.046\textwidth,trim=45.2mm 45.2mm 361.6mm 0,clip]{ret_1.jpg}\hspace{-1mm}
\includegraphics[height=.046\textwidth,trim=45.2mm 0 361.6mm 45.2mm,clip]{ret_1.jpg}
\includegraphics[height=.046\textwidth,trim=226mm 45.2mm 180.8mm 0,clip]{ret_1.jpg}\hspace{-1mm}
\includegraphics[height=.046\textwidth,trim=226mm 0 180.8mm 45.2mm,clip]{ret_1.jpg}
\includegraphics[height=.046\textwidth,trim=316.4mm 45.2mm 90.4mm 0,clip]{ret_1.jpg}\hspace{-1mm}
\includegraphics[height=.046\textwidth,trim=316.4mm 0 90.4mm 45.2mm,clip]{ret_1.jpg}
\includegraphics[height=.046\textwidth,trim=406.8mm 45.2mm 0 0,clip]{ret_1.jpg}\hspace{-1mm}
\includegraphics[height=.046\textwidth,trim=406.8mm 0 0 45.2mm,clip]{ret_1.jpg}
\\
\includegraphics[height=.046\textwidth,trim=0 22.6mm 452mm 22.6mm,clip]{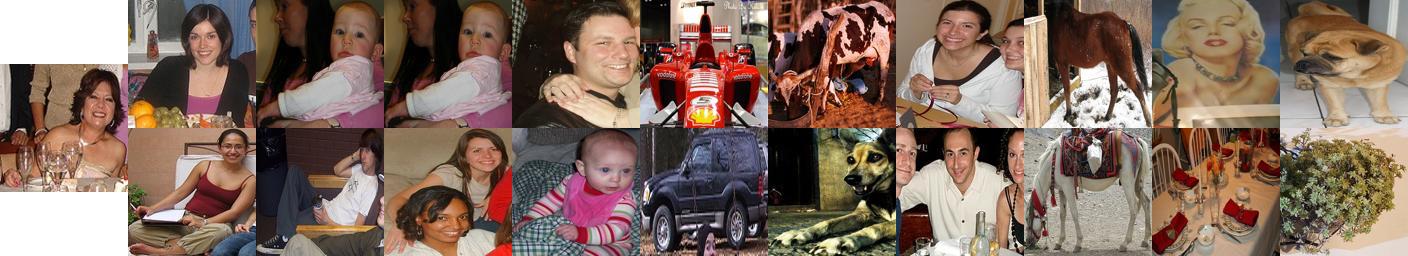}
\includegraphics[height=.046\textwidth,trim=135.6mm 45.2mm 271.2mm 0,clip]{ret_2.jpg}\hspace{-1mm}
\includegraphics[height=.046\textwidth,trim=135.6mm 0 271.2mm 45.2mm,clip]{ret_2.jpg}
\includegraphics[height=.046\textwidth,trim=45.2mm 45.2mm 361.6mm 0,clip]{ret_2.jpg}\hspace{-1mm}
\includegraphics[height=.046\textwidth,trim=45.2mm 0 361.6mm 45.2mm,clip]{ret_2.jpg}
\includegraphics[height=.046\textwidth,trim=226mm 45.2mm 180.8mm 0,clip]{ret_2.jpg}\hspace{-1mm}
\includegraphics[height=.046\textwidth,trim=226mm 0 180.8mm 45.2mm,clip]{ret_2.jpg}
\includegraphics[height=.046\textwidth,trim=316.4mm 45.2mm 90.4mm 0,clip]{ret_2.jpg}\hspace{-1mm}
\includegraphics[height=.046\textwidth,trim=316.4mm 0 90.4mm 45.2mm,clip]{ret_2.jpg}
\includegraphics[height=.046\textwidth,trim=406.8mm 45.2mm 0 0,clip]{ret_2.jpg}\hspace{-1mm}
\includegraphics[height=.046\textwidth,trim=406.8mm 0 0 45.2mm,clip]{ret_2.jpg}
\\
\includegraphics[height=.046\textwidth,trim=0 22.6mm 452mm 22.6mm,clip]{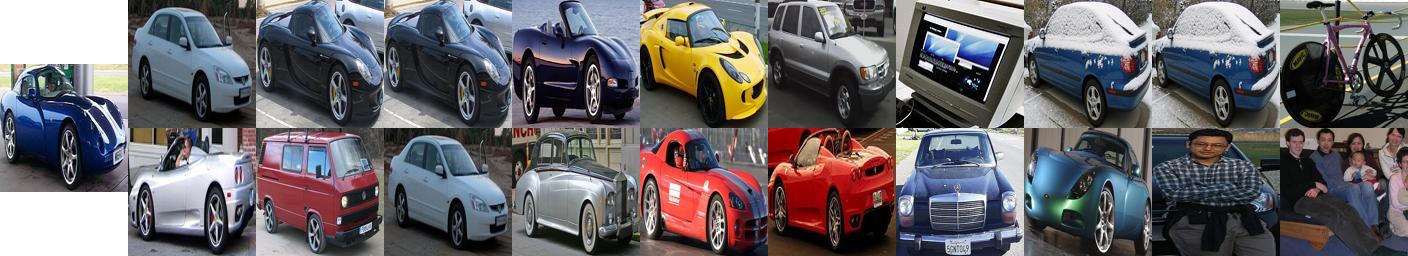}
\includegraphics[height=.046\textwidth,trim=135.6mm 45.2mm 271.2mm 0,clip]{ret_3.jpg}\hspace{-1mm}
\includegraphics[height=.046\textwidth,trim=135.6mm 0 271.2mm 45.2mm,clip]{ret_3.jpg}
\includegraphics[height=.046\textwidth,trim=45.2mm 45.2mm 361.6mm 0,clip]{ret_3.jpg}\hspace{-1mm}
\includegraphics[height=.046\textwidth,trim=45.2mm 0 361.6mm 45.2mm,clip]{ret_3.jpg}
\includegraphics[height=.046\textwidth,trim=226mm 45.2mm 180.8mm 0,clip]{ret_3.jpg}\hspace{-1mm}
\includegraphics[height=.046\textwidth,trim=226mm 0 180.8mm 45.2mm,clip]{ret_3.jpg}
\includegraphics[height=.046\textwidth,trim=316.4mm 45.2mm 90.4mm 0,clip]{ret_3.jpg}\hspace{-1mm}
\includegraphics[height=.046\textwidth,trim=316.4mm 0 90.4mm 45.2mm,clip]{ret_3.jpg}
\includegraphics[height=.046\textwidth,trim=406.8mm 45.2mm 0 0,clip]{ret_3.jpg}\hspace{-1mm}
\includegraphics[height=.046\textwidth,trim=406.8mm 0 0 45.2mm,clip]{ret_3.jpg}
\\
\includegraphics[height=.046\textwidth,trim=0 22.6mm 452mm 22.6mm,clip]{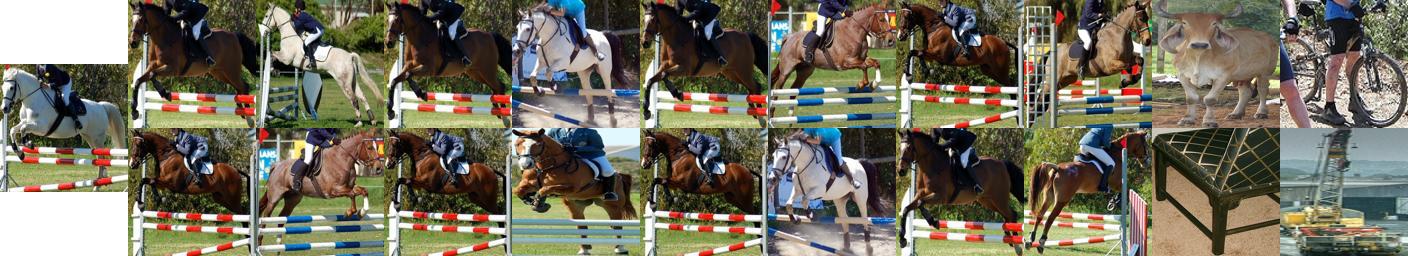}
\includegraphics[height=.046\textwidth,trim=135.6mm 45.2mm 271.2mm 0,clip]{ret_4.jpg}\hspace{-1mm}
\includegraphics[height=.046\textwidth,trim=135.6mm 0 271.2mm 45.2mm,clip]{ret_4.jpg}
\includegraphics[height=.046\textwidth,trim=45.2mm 45.2mm 361.6mm 0,clip]{ret_4.jpg}\hspace{-1mm}
\includegraphics[height=.046\textwidth,trim=45.2mm 0 361.6mm 45.2mm,clip]{ret_4.jpg}
\includegraphics[height=.046\textwidth,trim=226mm 45.2mm 180.8mm 0,clip]{ret_4.jpg}\hspace{-1mm}
\includegraphics[height=.046\textwidth,trim=226mm 0 180.8mm 45.2mm,clip]{ret_4.jpg}
\includegraphics[height=.046\textwidth,trim=316.4mm 45.2mm 90.4mm 0,clip]{ret_4.jpg}\hspace{-1mm}
\includegraphics[height=.046\textwidth,trim=316.4mm 0 90.4mm 45.2mm,clip]{ret_4.jpg}
\includegraphics[height=.046\textwidth,trim=406.8mm 45.2mm 0 0,clip]{ret_4.jpg}\hspace{-1mm}
\includegraphics[height=.046\textwidth,trim=406.8mm 0 0 45.2mm,clip]{ret_4.jpg}
\\
\includegraphics[height=.046\textwidth,trim=0 22.6mm 452mm 22.6mm,clip]{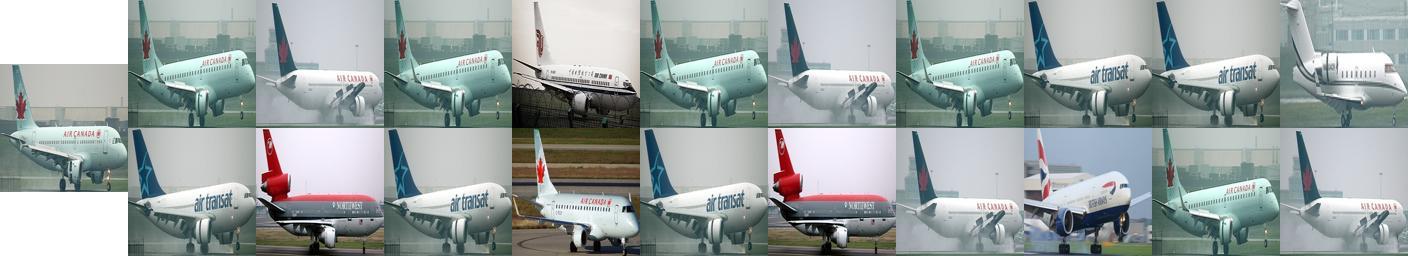}
\includegraphics[height=.046\textwidth,trim=135.6mm 45.2mm 271.2mm 0,clip]{ret_5.jpg}\hspace{-1mm}
\includegraphics[height=.046\textwidth,trim=135.6mm 0 271.2mm 45.2mm,clip]{ret_5.jpg}
\includegraphics[height=.046\textwidth,trim=45.2mm 45.2mm 361.6mm 0,clip]{ret_5.jpg}\hspace{-1mm}
\includegraphics[height=.046\textwidth,trim=45.2mm 0 361.6mm 45.2mm,clip]{ret_5.jpg}
\includegraphics[height=.046\textwidth,trim=226mm 45.2mm 180.8mm 0,clip]{ret_5.jpg}\hspace{-1mm}
\includegraphics[height=.046\textwidth,trim=226mm 0 180.8mm 45.2mm,clip]{ret_5.jpg}
\includegraphics[height=.046\textwidth,trim=316.4mm 45.2mm 90.4mm 0,clip]{ret_5.jpg}\hspace{-1mm}
\includegraphics[height=.046\textwidth,trim=316.4mm 0 90.4mm 45.2mm,clip]{ret_5.jpg}
\includegraphics[height=.046\textwidth,trim=406.8mm 45.2mm 0 0,clip]{ret_5.jpg}\hspace{-1mm}
\includegraphics[height=.046\textwidth,trim=406.8mm 0 0 45.2mm,clip]{ret_5.jpg}
\\
\includegraphics[height=.046\textwidth,trim=0 22.6mm 452mm 22.6mm,clip]{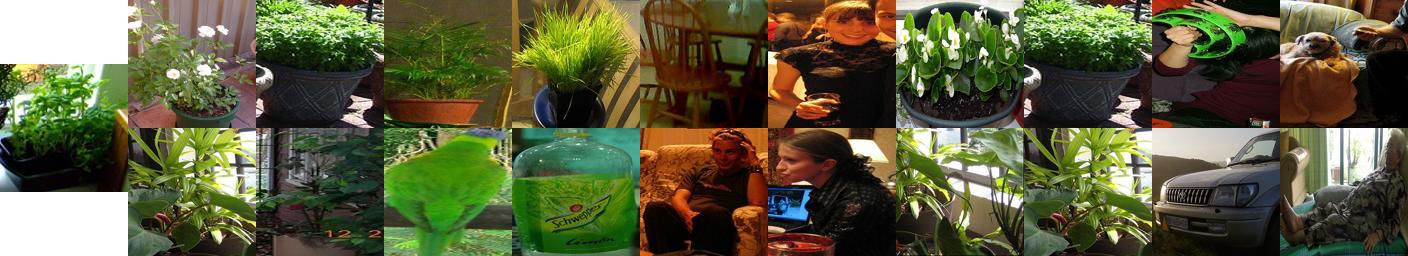}
\includegraphics[height=.046\textwidth,trim=135.6mm 45.2mm 271.2mm 0,clip]{ret_6.jpg}\hspace{-1mm}
\includegraphics[height=.046\textwidth,trim=135.6mm 0 271.2mm 45.2mm,clip]{ret_6.jpg}
\includegraphics[height=.046\textwidth,trim=45.2mm 45.2mm 361.6mm 0,clip]{ret_6.jpg}\hspace{-1mm}
\includegraphics[height=.046\textwidth,trim=45.2mm 0 361.6mm 45.2mm,clip]{ret_6.jpg}
\includegraphics[height=.046\textwidth,trim=226mm 45.2mm 180.8mm 0,clip]{ret_6.jpg}\hspace{-1mm}
\includegraphics[height=.046\textwidth,trim=226mm 0 180.8mm 45.2mm,clip]{ret_6.jpg}
\includegraphics[height=.046\textwidth,trim=316.4mm 45.2mm 90.4mm 0,clip]{ret_6.jpg}\hspace{-1mm}
\includegraphics[height=.046\textwidth,trim=316.4mm 0 90.4mm 45.2mm,clip]{ret_6.jpg}
\includegraphics[height=.046\textwidth,trim=406.8mm 45.2mm 0 0,clip]{ret_6.jpg}\hspace{-1mm}
\includegraphics[height=.046\textwidth,trim=406.8mm 0 0 45.2mm,clip]{ret_6.jpg}
\\
\includegraphics[height=.046\textwidth,trim=0 22.6mm 452mm 22.6mm,clip]{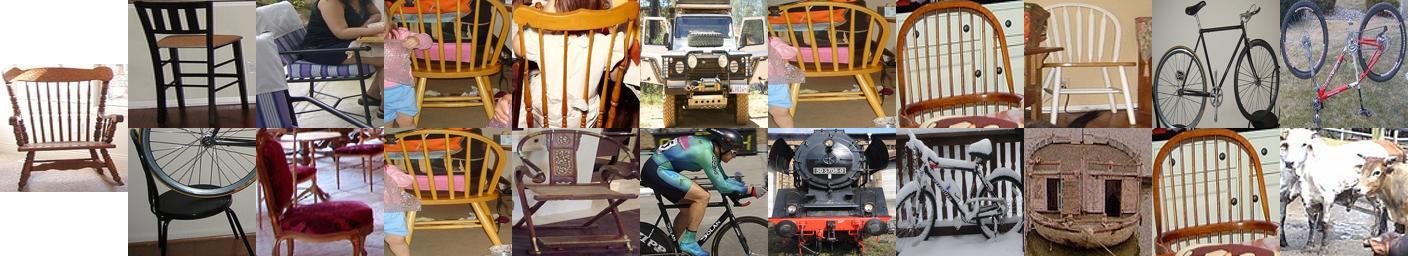}
\includegraphics[height=.046\textwidth,trim=135.6mm 45.2mm 271.2mm 0,clip]{ret_10.jpg}\hspace{-1mm}
\includegraphics[height=.046\textwidth,trim=135.6mm 0 271.2mm 45.2mm,clip]{ret_10.jpg}
\includegraphics[height=.046\textwidth,trim=45.2mm 45.2mm 361.6mm 0,clip]{ret_10.jpg}\hspace{-1mm}
\includegraphics[height=.046\textwidth,trim=45.2mm 0 361.6mm 45.2mm,clip]{ret_10.jpg}
\includegraphics[height=.046\textwidth,trim=226mm 45.2mm 180.8mm 0,clip]{ret_10.jpg}\hspace{-1mm}
\includegraphics[height=.046\textwidth,trim=226mm 0 180.8mm 45.2mm,clip]{ret_10.jpg}
\includegraphics[height=.046\textwidth,trim=316.4mm 45.2mm 90.4mm 0,clip]{ret_10.jpg}\hspace{-1mm}
\includegraphics[height=.046\textwidth,trim=316.4mm 0 90.4mm 45.2mm,clip]{ret_10.jpg}
\includegraphics[height=.046\textwidth,trim=406.8mm 45.2mm 0 0,clip]{ret_10.jpg}\hspace{-1mm}
\includegraphics[height=.046\textwidth,trim=406.8mm 0 0 45.2mm,clip]{ret_10.jpg}
\\
\includegraphics[height=.046\textwidth,trim=0 22.6mm 452mm 22.6mm,clip]{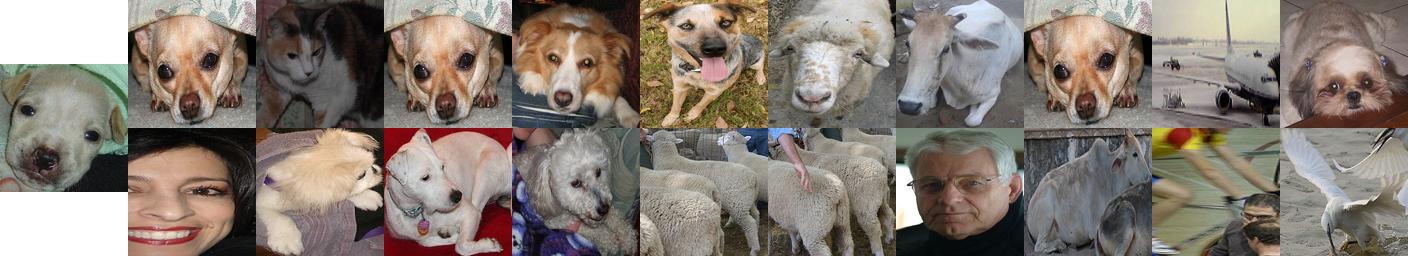}
\includegraphics[height=.046\textwidth,trim=135.6mm 45.2mm 271.2mm 0,clip]{ret_7.jpg}\hspace{-1mm}
\includegraphics[height=.046\textwidth,trim=135.6mm 0 271.2mm 45.2mm,clip]{ret_7.jpg}
\includegraphics[height=.046\textwidth,trim=45.2mm 45.2mm 361.6mm 0,clip]{ret_7.jpg}\hspace{-1mm}
\includegraphics[height=.046\textwidth,trim=45.2mm 0 361.6mm 45.2mm,clip]{ret_7.jpg}
\includegraphics[height=.046\textwidth,trim=226mm 45.2mm 180.8mm 0,clip]{ret_7.jpg}\hspace{-1mm}
\includegraphics[height=.046\textwidth,trim=226mm 0 180.8mm 45.2mm,clip]{ret_7.jpg}
\includegraphics[height=.046\textwidth,trim=316.4mm 45.2mm 90.4mm 0,clip]{ret_7.jpg}\hspace{-1mm}
\includegraphics[height=.046\textwidth,trim=316.4mm 0 90.4mm 45.2mm,clip]{ret_7.jpg}
\includegraphics[height=.046\textwidth,trim=406.8mm 45.2mm 0 0,clip]{ret_7.jpg}\hspace{-1mm}
\includegraphics[height=.046\textwidth,trim=406.8mm 0 0 45.2mm,clip]{ret_7.jpg}
\\
\includegraphics[height=.046\textwidth,trim=0 22.6mm 452mm 22.6mm,clip]{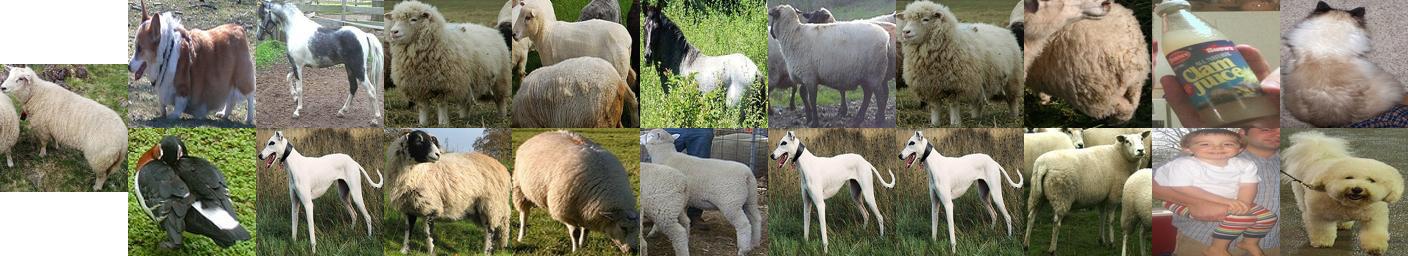}
\includegraphics[height=.046\textwidth,trim=135.6mm 45.2mm 271.2mm 0,clip]{ret_8.jpg}\hspace{-1mm}
\includegraphics[height=.046\textwidth,trim=135.6mm 0 271.2mm 45.2mm,clip]{ret_8.jpg}
\includegraphics[height=.046\textwidth,trim=45.2mm 45.2mm 361.6mm 0,clip]{ret_8.jpg}\hspace{-1mm}
\includegraphics[height=.046\textwidth,trim=45.2mm 0 361.6mm 45.2mm,clip]{ret_8.jpg}
\includegraphics[height=.046\textwidth,trim=226mm 45.2mm 180.8mm 0,clip]{ret_8.jpg}\hspace{-1mm}
\includegraphics[height=.046\textwidth,trim=226mm 0 180.8mm 45.2mm,clip]{ret_8.jpg}
\includegraphics[height=.046\textwidth,trim=316.4mm 45.2mm 90.4mm 0,clip]{ret_8.jpg}\hspace{-1mm}
\includegraphics[height=.046\textwidth,trim=316.4mm 0 90.4mm 45.2mm,clip]{ret_8.jpg}
\includegraphics[height=.046\textwidth,trim=406.8mm 45.2mm 0 0,clip]{ret_8.jpg}\hspace{-1mm}
\includegraphics[height=.046\textwidth,trim=406.8mm 0 0 45.2mm,clip]{ret_8.jpg}
\\
\includegraphics[height=.046\textwidth,trim=0 22.6mm 452mm 22.6mm,clip]{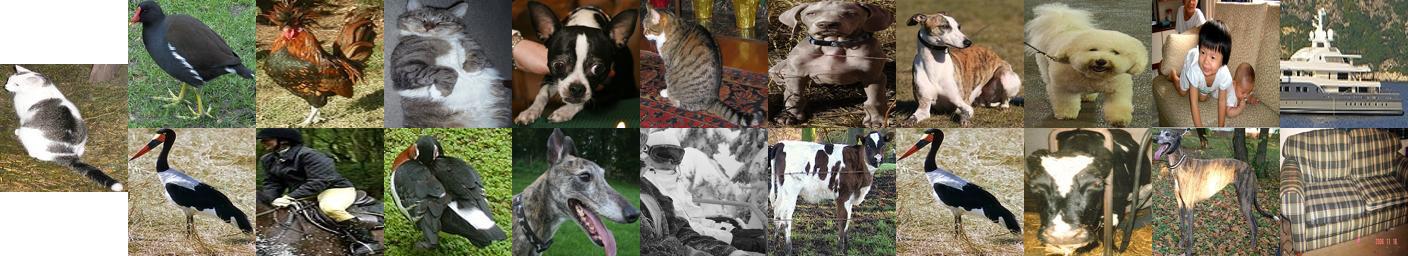}
\includegraphics[height=.046\textwidth,trim=135.6mm 45.2mm 271.2mm 0,clip]{ret_9.jpg}\hspace{-1mm}
\includegraphics[height=.046\textwidth,trim=135.6mm 0 271.2mm 45.2mm,clip]{ret_9.jpg}
\includegraphics[height=.046\textwidth,trim=45.2mm 45.2mm 361.6mm 0,clip]{ret_9.jpg}\hspace{-1mm}
\includegraphics[height=.046\textwidth,trim=45.2mm 0 361.6mm 45.2mm,clip]{ret_9.jpg}
\includegraphics[height=.046\textwidth,trim=226mm 45.2mm 180.8mm 0,clip]{ret_9.jpg}\hspace{-1mm}
\includegraphics[height=.046\textwidth,trim=226mm 0 180.8mm 45.2mm,clip]{ret_9.jpg}
\includegraphics[height=.046\textwidth,trim=316.4mm 45.2mm 90.4mm 0,clip]{ret_9.jpg}\hspace{-1mm}
\includegraphics[height=.046\textwidth,trim=316.4mm 0 90.4mm 45.2mm,clip]{ret_9.jpg}
\includegraphics[height=.046\textwidth,trim=406.8mm 45.2mm 0 0,clip]{ret_9.jpg}\hspace{-1mm}
\includegraphics[height=.046\textwidth,trim=406.8mm 0 0 45.2mm,clip]{ret_9.jpg}
\\
\begin{tabular}{cccccc}
(a)&\hspace{9mm}(b)&\hspace{19mm}(c)&\hspace{18mm}(d)&\hspace{18mm}(e)
\hspace{18mm}(f)&\hspace{10mm}
\end{tabular}
 \caption{Image retrieval (qualitative evaluation). (a) query images; (b) top-4 matches with AlexNet; (c) top-4 matches with the CFN  trained without blocking chromatic aberration; (d) top-4 matches with Doersch~\etal \cite{Carl2015}; (e) top-4 matches with Wang and Gupta \cite{Gupta15}; (f) top-4 matches with AlexNet with random weights.} \label{fig:retrievalquali}
\end{center}
\end{figure}

\begin{figure}[t!]
\begin{center}
\includegraphics[width=.7\textwidth,height=.25\textheight, trim=15mm 75mm 25mm 85mm,clip]{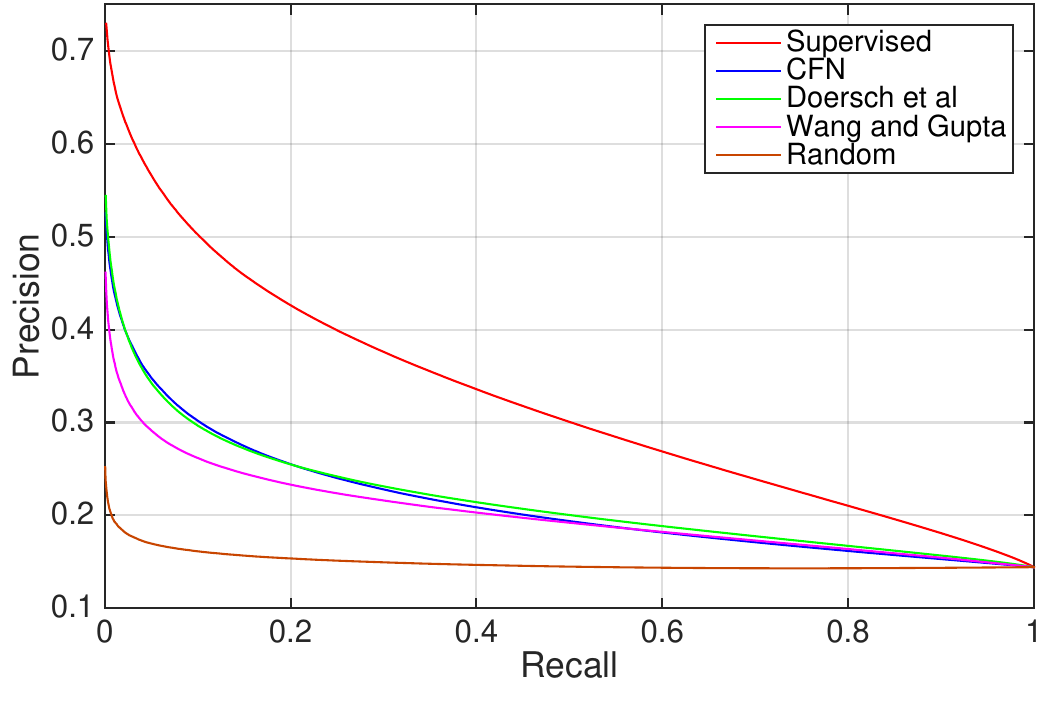}
\caption{Image retrieval (quantitative evaluation). We compare the precision-recall for image retrieval on the PASCAL VOC 2007. The ranking of the retrieved images is based on the inner products between normalized features extracted from a pre-trained AlexNet, the CFN, Doersch \etal \cite{Carl2015}, Wang and Gupta \cite{Gupta15} and from AlexNet with random weights. The performance of CFN and \cite{Carl2015} are very similar when using this simple ranking metric. When the metric is instead learned with two fully connected layers, then we see that CFN features yield a clearly higher performance than all other features from self-supervised learning methods (see Table~\ref{tab:classification}).} \label{fig:retrievalquanti}
\end{center}
\end{figure}

We also evaluate the features qualitatively (see Fig.~\ref{fig:retrievalquali}) and quantitatively (see Fig.~\ref{fig:retrievalquanti}) for image retrieval with a simple image ranking.

We find the nearest neighbors (NN) of \texttt{pool5} features using the bounding boxes of the PASCAL VOC 2007 \emph{test} set as query and bounding boxes of the \emph{trainval} set as the retrieval entries. We discard bounding boxes with fewer than $10$K pixels inside. In Fig.~\ref{fig:retrievalquali} we show some examples of image retrievals (top-4) obtained by ranking the images based on the inner product between normalized  features of a query image and normalized features of the retrieval set. We can see that the features of the CFN are very sensitive to objects with similar shape and often these are within the same category. In Fig.~\ref{fig:retrievalquanti} we compare CFN with the pre-trained AlexNet, \cite{Carl2015}, \cite{Gupta15}, and AlexNet with random weights. The precision-recall plots show that \cite{Carl2015} and CFN features perform equally well. However, the real potential of CFN features is demonstrated when the feature metric is learned. In Table~\ref{tab:classification} we can see how CFN features surpass other features trained in an unsupervised way by a good margin. In that test the dataset (ImageNet) is more challenging because there are more categories and the bounding box is not used.

\clearpage
\section{Conclusions}

We have introduced the \emph{context-free} network (CFN), a CNN whose features can be easily transferred between detection/classification and Jigsaw puzzle reassembly tasks. The network is trained in an unsupervised manner by using the Jigsaw puzzle as a \emph{pretext} task. We have built a training scheme that generates, on average, $69$ puzzles for $1.3$M images and converges in only $2.5$ days. The key idea is that by solving Jigsaw puzzles the CFN learns to identify each tile as an object part and how parts are assembled in an object. The learned features are evaluated on both classification and detection and the experiments show that we outperform the previous state of the art. More importantly, the performance of these features is closing the gap with those learned in a supervised manner. We believe that there is a lot of untapped potential in self-supervised learning and in the future it will provide a valid alternative to costly human annotation.

\bibliographystyle{splncs03}

\bibliography{puzzle_learning}

\end{document}